\documentclass[runningheads]{llncs}
\usepackage{graphicx}
\usepackage{subfigure}
\usepackage{amssymb}
\usepackage{amsmath}
\usepackage{multirow}
\usepackage{booktabs}
\usepackage{bigstrut}
\usepackage[misc]{ifsym}
\usepackage{makecell}
 \usepackage{bigstrut}
\usepackage[table]{xcolor}
\begin{document}

\title{Weakly Supervised Online Action Detection for Infant General Movements}
\authorrunning{T. Luo et al.}

\author{Tongyi Luo\inst{1} \and
Jia Xiao\inst{1} \and
Chuncao Zhang\inst{3} \and Siheng Chen\inst{1} \and Yuan Tian\inst{3} \and Guangjun Yu\inst{3} \and Kang Dang\inst{2 \thanks{Co-Corresponding Author}} \and Xiaowei Ding\inst{1, 2\footnotemark[1]} \textsuperscript{(\Letter)}}
% 1{Luo, Tongyi} 2{Xiao, jia} 3{Zhang, Chuncao} 4{Chen, Siheng} 5{Tian, Yuan}
% 6{Yu, Guangjun} 7{Dang, Kang} 8{Ding, Xiaowei}

\institute{Shanghai Jiao Tong University, Shanghai, China \and
VoxelCloud, Inc., Los Angeles, USA \and
Department of Health Management, Shanghai Children's Hospital, Shanghai, China\\
\email{dingxiaowei@sjtu.edu.cn}}
\maketitle              % typeset the header of the contribution
\begin{abstract}
To make the earlier medical intervention of infants' cerebral palsy (CP), early diagnosis of brain damage is critical. Although general movements assessment(GMA) has shown promising results in early CP detection, it is laborious. Most existing works take videos as input to make fidgety movements(FMs) classification for the GMA automation. Those methods require a complete observation of videos and can not localize video frames containing normal FMs. Therefore we propose a novel approach named WO-GMA to perform FMs localization in the weakly supervised online setting. Infant body keypoints are first extracted as the inputs to WO-GMA. Then WO-GMA performs local spatio-temporal extraction followed by two network branches to generate pseudo clip labels and model online actions. With the clip-level pseudo labels, the action modeling branch learns to detect FMs in an online fashion. Experimental results on a dataset with 757 videos of different infants show that WO-GMA can get state-of-the-art video-level classification and clip-level detection results. Moreover, only the first 20\% duration of the video is needed to get classification results as good as fully observed, implying a significantly shortened FMs diagnosis time. Code is available at: \url{https://github.com/scofiedluo/WO-GMA}.
\keywords{Online action detection \and weakly supervised \and general movements assessment \and fidgety movements(FMs).}
\end{abstract}
\section{Introduction} 
Clinical and public health problems of surviving high-risk infants are common worldwide. Taking preterm infants with the highest proportion of high-risk infants as an example, they may face various complications that affect the quality of life, such as delayed growth in language, cognition, motor, intelligence, and even cause cerebral palsy(CP) \cite{malcolm2015beyond}. Diagnoses of CP are critical for early intervention of such high-risk preterm infants. 

Studies indicate that general movements assessment(GMA) with high sensitivity ($98\%$) and specificity ($91\%$) is the most cost-effective and accurate tool for early diagnoses of CP \cite{herskind2015early}. Fidgety movements(FMs) is an important stage of general movements(GMs). Normal FMs(F+) are smoothly circular movements involving the whole body, including the neck, trunk, and limbs. These movements are small in amplitude, moderate in speed, and variable in all directions \cite{einspieler1997qualitative,prechtl1986developmental}. The absence or sporadic occurrence of FMs(F-) is a strong indicator for infants' CP risk \cite{einspieler2016fidgety}. Qualified assessors usually watch the videos of infants to identify the absence or sporadic occurrence of FMs.

Though GMA is highly accurate, there is a great shortage of qualified assessors, and the assessment is time-consuming. Many works use machine learning or deep learning methods to make GMA automated. Generally, according to the sensors type, automated GMA can be categorized into vision-sensor-based and motion-sensor-based \cite{irshad2020ai}. Since 2D camera data is easier to collect, we focus on vision-based methods. RGB frames are directly processed by VGG followed by LSTM to capture temporal information in \cite{schmidt2019general}. However, RGB frames data contains lots of irrelevant noise in GMA scenes, such as illumination, background, and camera properties. Most vision-based works \cite{nguyen2021spatio,chambers2020computer,wu2021automatically,mccay2021towards} claim that extracting body keypoints from video data followed by keypoints motion feature analyzer is more robust. Existing works focus on video classification after fully observing the video, leaving two critical problems unaddressed. First, for the real-world application of automated GMA, deep learning methods need greater interpretability. Since unhealthy infants would not exhibit normal FMs for a long time, we can assess infants by localizing when the normal FMs occur. Besides, if the assessment can be completed by partially observing the video, the record time(diagnosis time) can be shortened.

This paper addresses the above two key challenges by proposing a framework in the weakly supervised online action detection setting named WO-GMA. Most online action detection methods \cite{geest2016online,xu2019temporal,xu2021long,eun2020learning} rely on frame-level annotations of action boundaries for training. Annotating action boundaries may involve ambiguous decisions and is laborious, especially for FMs mixed with other movements; hence weakly supervised methods are preferred for FMs detection. Many weakly supervised action detection methods \cite{gao2021woad,paul2018w,zhang2021cola} utilize multiple instance learning \cite{zhou2004multi} or contrastive learning \cite{gutmann2010noise} to train models with video-level labels. Following previous works, we first extract infants' 2D poses from videos as input of WO-GMA. The pipeline of our method is as follows. WO-GMA contains one local spatio-temporal extraction module followed by two branches to generate pseudo labels and model online action. The local module uses a 3D graph convolution network \cite{liu2020disentangling} to capture complex spatio-temporal features based on the extracted infant poses. Supervised by video-level labels, clip-level pseudo labels generating branch mines temporal labels by mixing local feature and long-range information. The online action modeling branch utilizes the generated pseudo labels to conduct clip-level action detection without future information.

\textbf{Contributions}: (1) We are the first to develop an online action detection method on this task and report frame-level recognition results. (2) We validate WO-GMA on our dataset with 757 videos of different infants. Experiments show that the video-level FMs prediction of our method outperforms existing automated GMA models. (3) Experiments demonstrate that we can obtain accurate video level results when only the first $20\%$ of full video frames are used, implying a shortened FMs diagnosis time.

\section{Methodology}

\begin{figure}[htbp]
\includegraphics[width=\textwidth]{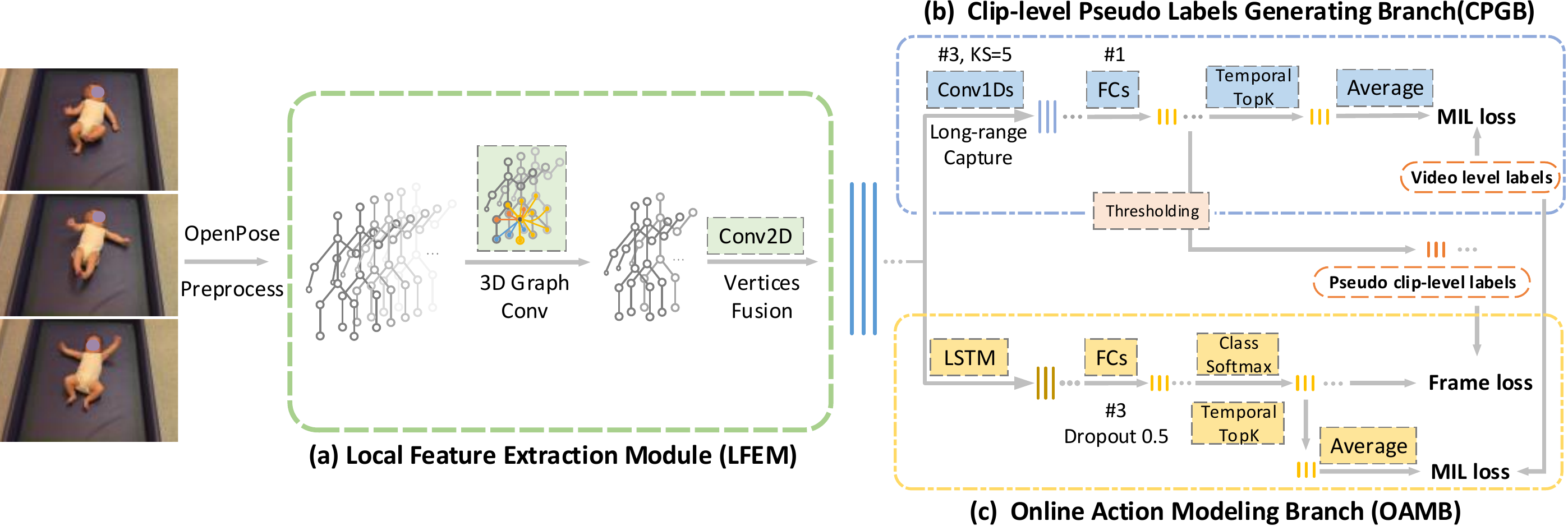}
\caption{Overall framework architecture of WO-GMA. Kernel Size(KS), Layers(\#)} 
\label{fig1}
\end{figure}

We use a sequence of 2D keypoints estimated from the RGB video frames as our network input. Figure~\ref{fig1}) shows the network architecture of our proposed skeleton-based weakly supervised online action detection model. It consists of three main components. First, a local feature extraction module (LFEM), containing a spatio-temporal graph network followed by joints fusing, is used to extract complex local features from the skeletons. Then, we capture bidirectional long-range information with a clip-level pseudo labels generating branch(CPGB) supervised by video-level labels. Third, the online action modeling branch(OAMB) supervised by the generated pseudo labels is used to detect action without future information. These components will be detailed in this section.

\subsection{Local Feature Extraction Module}
Human pose keypoints named skeleton is usually denoted as $\mathcal{G}=(\mathcal{V},\mathcal{E})$, where nodes set $\mathcal{V}=\{v_i|1 \leq i \leq N\}$ represents joints, and edges set $\mathcal{E}=\{e_i|1\leq i \leq E\}$ represents connectivity between joints. Formally, we use adjacency matrix $\mathbf{A} \in \mathbb{R}^{N \times N}$ to denote the edges set where $A_{i,j} = 1$ if there is an edge between $v_i$ and $v_j$, otherwise 0. Assume that we are given a skeleton sequence $\mathcal{X} = \{\mathbf{x}_{t,n} \in \mathbb{R}^C | t,n \in \mathbb{Z}, 1 \leq t \leq T, 1 \leq n \leq N\}$, where $T$ is the sequence length, and $C$ denotes the feature vector dimension for one joint. Then the feature of this sequence can be written as $\mathbf{X} \in \mathbb{R}^{T\times N \times C}$.

As detailed in section 1, FMs have complex movement patterns. To well modeling the local spatio-temporal information, we need to capture the relation between different joints in one frame and the variations of joints over time. Multi-scale graph convolution is often used to fuse long-range relations in one graph \cite{li2019actional}. MS-G3D \cite{liu2020disentangling} utilized cross-spacetime skip connections to construct a spatio-temporal subgraph and proposed a disentangled multi-scale graph convolution to model human action dynamics. We extend the feature extractor in MS-G3D with vertices fusing to capture the local spatio-temporal information of FMs.

In detail, we first split the pose sequence $\mathcal{X}$ into clip-level set $\{C_1,C2,\cdots, C_L\}$ with a sliding window of temporal size $\tau$ and stride size $s$, where $C_1 \cup C_2 \cup \cdots \cup C_L = \mathcal{X}$. Then a spatio-temporal graph $(\mathcal{V}_{\tau},\mathcal{E}_{\tau})$ is constructed by tiling $\mathbf{A}$ into a block adjacency matrix $\mathbf{A}_{\tau} \in \mathbb{R}^{\tau N \times \tau N}$. A multi-scale graph convolution is applied to each clip feature tensor  $\mathbf{X}^{in}_{(\tau)} \in \mathbb{R}^{1 \times \tau N \times C_{in}}$ as
\begin{equation}
    \mathbf{X}^{out}_{(\tau)}=\sigma\left(\sum_{m=0}^{M} \tilde{\mathbf{D}}_{(\tau, m)}^{-\frac{1}{2}} \tilde{\mathbf{A}}_{(\tau, m)} \tilde{\mathbf{D}}_{(\tau, m)}^{-\frac{1}{2}}\mathbf{X}^{in}_{(\tau)} \mathbf{\Theta}_{(m)}\right), 
\end{equation}
where $\mathbf{X}^{out}_{(\tau)} \in \mathbb{R}^{1 \times \tau N \times C_{out}}$ is output feature, $\sigma$ is activation function, $M  \in \mathbf{R}$ is the number of scales to aggregate.   $\tilde{\mathbf{D}}_{(\tau, m)}$ is the corresponding diagonal degree matrix of disentangled $\tilde{\mathbf{A}}_{(\tau, m)}$. $\mathbf{\Theta}_{(m)} \in \mathbb{R}^{C_{in}\times C_{out}}$ denotes the learnable parameter matrix.

For each clip, $\mathbf{X}^{out}_{(\tau)}$ will be collapsed into one skeleton by a 3D convolution operator with kernel size $(1,\tau,1)$ to get a feature tensor $\mathbf{X}^{out} \in \mathbb{R}^{1 \times N \times C_{out}}$. Then each joint in $\mathbf{X}^{out}$ has got rich information from their neighborhood. To extract more complex local spatio-temporal fused feature, we will aggregate joints information as $\mathbf{f}_{i} = \mbox{agg}(\mathbf{X}_{i}^{out})$, 
where subscript $i$ denotes the index of clip. We use 2D convolution as aggregator followed with ReLU in this paper. Then we get a feature sequences as $[\mathbf{f}_1,\mathbf{f}_2,\cdots,\mathbf{f}_L]$. All the clips share the same parameter in this module.

\subsection{Clip-level Pseudo Labels Generating Branch}
For inference of weakly supervised online action detection, only accumulated history information is available. We introduce a branch to generate pseudo labels with future information and long-range information during training. Since infants' FMs are continual and hard to distinguish from other mixed movements \cite{einspieler1997qualitative}, combination future information with long time receptive field is helpful for better classification and detection.

Take the features $\mathbf{F}^{0} = [\mathbf{f}_1,\mathbf{f}_2,\cdots,\mathbf{f}_L] \in \mathbb{R}^{L \times C_{out}}$ extracted from LFEM as input, we use the 1D temporal convolutions followed by ReLU to aggregate long-range temporal information from neighbour clips features, i.e. $\mathbf{F}^{i+1} = \mbox{Relu}(\mbox{Conv1D}(\mathbf{F}^{i}))$.
The output feature of the last 1D convolution layer is $\mathbf{F}_{out}$. Then for each clip, fully connected layers are used to get the action scores $\mathbf{S} = [\mathbf{s}_1,\mathbf{s}_2, \cdots, \mathbf{s}_L]$, where $\mathbf{s}_i \in \mathbb{R}^{n_c}$ is the score of $n_c$ actions of clip $i$.

Multiple Instance Learning Loss(MILL) is widely used to get accurate actions scores \cite{paul2018w,gao2021woad}. In this paper, we consider the entire video clip-level sequence set $\{C_1,C2,\cdots, C_L\}$ as a bag of instances, where each instance denotes one clip. To use the video-level labels, we compute video-level scores with Top-K strategy for each action class. That is $s^{c} = \frac{1}{K}\sum_{j \in \mathcal{M}}s^{c}_{j}$, where $\mathcal{M}$ is the Top-K indices set of clips over class $c$, and $K$ is got by $\max (1,\left\lfloor\frac{L_{}}{\kappa}\right\rfloor)$ with hyper parameter $\kappa$. Then, to obtain the action class probabilities $\mathcal{P}$, sigmoid or softmax (multi-class dataset) is applied to the video-level scores. The MILL is computed as 
\begin{equation}
    \mathcal{L}_{MIL}^{p} = - \sum_{c=1}^{n_c} y_c \log (p_c),
    \label{mill1}
\end{equation}
where $y_c$ and $p_c$ is the ground truth label and prediction probability of class $c$.
Supervised by $\mathcal{L}_{MIL}^{p}$, the network can learn clip-level scores which will be used to generate clip-level pseudo labels with future information by a two-stage threshold strategy \cite{gao2021woad}. First, action class will be discarded if video-level score is less than a threshold $\theta_{class}$. Then, for the remaining action classes, a second threshold, $\theta_{score}$, is applied on clip-level action scores $\mathbf{S}$ to get pseudo labels $L_{P}$. After that, the video-level ground truth labels are used to filter out wrong pseudo labels.

\subsection{Online Action Modeling Branch}
As shown above, CPGB can utilize future information during training. However, future information is not available during inference. LSTM is used to accumulate the historical temporal information in this branch.

Given one clip-level feature $\mathbf{f}_i$, previous hidden state $\mathbf{h}_{i-1}$ and cell state $\mathbf{c}_{i-1}$ as input, updated states is got by $\mathbf{h}_{i},\mathbf{c}_{i}= \mbox{LSTM} (\mathbf{h}_{i-1}, \mathbf{c}_{i-1}, \mathbf{f}_i)$.
For $i\mbox{th}$ clip, online action score is computed as
$\mathbf{a}_{i} =\operatorname{softmax}\left(\mathbf{W}_{a}^{\top} \mathbf{h}_{i}\right)$, 
where $\mathbf{a}_{i} \in \mathbb{R}^{n_c + 1}$ is the action scores including a background class. Then, cross entropy loss is applied to $\mathbf{a}_{i}$ over all clips with pseudo labels $L_P$ to get frame loss, i.e.
\begin{equation}
    \mathcal{L}_{FML} = -\frac{1}{L} \sum_{i=1}^{L} \sum_{c=0}^{n_c} L_{P,i}^c \log \mathbf{a}_i^c
\end{equation}
To further utilize the ground truth video level information, another MILL is used in this branch. As shown in Figure~\ref{fig1}, we use the same top-K strategy as CPGB to get video-level scores for each action class. Here, $\mathcal{L}_{MIL}^{o}$ is used to denote MILL in this branch.

\subsection{Training and Inference}
In the training stage, two branches and the local feature extraction module are jointly optimized by $\mathcal{L}_{MIL}^{p} + \mathcal{L}_{FML} + \mathcal{L}_{MIL}^{o}$. During inference, the future information is not available for online action detection tasks, so we only use the feature extraction module and online action modeling branch.

\section{Experiments}

\textbf{Dataset}. For this study, ethical approval was provided by the Ethics Committee of Children's Hospital of Shanghai (Review number: 2021RY053-E01). Written
informed consent was obtained from the parents/legal guardians in accordance with the Declaration of Helsinki. Our original dataset provided by three hospitals contains 792 videos. Two certified observers, who had got the GMs Trust qualification, made the video-level GMs classification. In case of disagreement, the third certified observer re-assessed the video. The eligible participants were high-risk (premature birth, low birth weight, suspected or brain injury, birth with chronic disease, genetic or genetic metabolic disease). Video recording was conducted according to standard protocol \cite{einspieler1997qualitative}. After deleting the repetitive videos, the dataset remains 757 different videos of 757 different infants at around 46$\sim$70 weeks gestational age(average: 55 weeks), including 353 F- videos and 404 F+ videos, 434 male and 323 female infants. The resolution of 678 videos is 1920$\times$1080, others are 1280$\times$720, 1440$\times$1080, 960$\times$540, 720$\times$576. The average number of frames is 7787(590$\sim$16322), and the average time duration is 307s(24$\sim$653s). We split the training and test sets with a ratio of 8:2, which ensures videos with different labels and of different hospitals are divided evenly.

\textbf{Evaluation Metrics}. For video-level performance, we report the classification accuracy, F1-score, and Area Under Curve(AUC). For detection results, following previous works \cite{paul2018w,zhang2021cola}, we use standard evaluation protocol by reporting mean Average Precision (mAP) values under different intersection over union (IoU) thresholds. 

\textbf{Implementation details}.
OpenPose \cite{cao2017realtime} is used to extract skeletons with 18 joints from videos and skeletons preprocessing method in \cite{chambers2020computer} is adopted. For convenience, we truncate the first 6000 frames for the skeleton sequences more than 6000 frames; otherwise, pad 0 at the end of each skeleton sequence to 6000 frames.
We implemented our WO-GMA in PyTorch, and performed experiments on a system with Nvidia 3090 GPUs. We train our network with 100 epochs by using Adam \cite{kingma2014adam} with learning rate $5\times10^{-5}$ and weight decay $5\times10^{-4}$. The window size $\tau$ and stride size $s$ are set to be 20. And the parameter $\kappa$ in the Top-K strategy of both branches is 8. Video threshold and frame threshold in pseudo labels generating are 0.4 and 0.3 respectively. The dimension of hidden state $\mathbf{h}_{i}$ in LSTM is 1024. Since we only focus on F+, the class number $n_c=1$.

\subsection{Main Results}

\begin{table}[t]
  \centering
  \caption{Comparison with other video-level classification works and action detection works on our datasets. Here fusion is the concatenation of image and optical flow features among channel dimensions. The skeleton-based video-level performances are evaluated with 5-fold cross-validation. The proposed WO-GMA outperforms previous works by a lot margin.}
    \begin{tabular}{lccccrcccccc}
    \toprule
    \multirow{2}[4]{*}{Input} & \multirow{2}[4]{*}{Method} & \multicolumn{3}{c}{Video level} &       & \multicolumn{6}{c}{Detection -- mAP@IoU(\%)} \\
\cmidrule{3-5}\cmidrule{7-12}          &       & Accuracy & F1    & AUC   &       & 0.1   & 0.2   & 0.3   & 0.4   & 0.5   & mean \\
\cmidrule{1-5}\cmidrule{7-12}    \multirow{2}[2]{*}{Image} & WOAD\cite{gao2021woad}  & 53.2  & 69.3  & 48    &       & 2.7  & 1.5  & 1.5  & 0.3  & 0.3  & 1.3 \\
          & W-TALC\cite{paul2018w} & 52.6  & 42.5  & 49.4  &       & 2.1  & 1.00  & 0.2  & 0.0     & 0.0     & 0.7 \\
\cmidrule{1-5}\cmidrule{7-12}    \multirow{2}[2]{*}{\shortstack{Optical\\ flow}} & WOAD\cite{gao2021woad}  & 54.5  & 63.5  & 51    &       & 11.5  & 7.8  & 5.0  & 3.2  & 1.9  & 5.9 \\
          & W-TALC\cite{paul2018w} & 57.8  & 67.9  & 55.9  &       & 5.6  & 1.1  & 0.1  & 0.0  & 0.0     & 1.4 \\
\cmidrule{1-5}\cmidrule{7-12}    \multirow{2}[2]{*}{Fusion} & WOAD\cite{gao2021woad}  & 55.2  & 69.9  & 49.4  &       & 11.5  & 8.7  & 5.5  & 3.9   & 1.7  & 6.3 \\
          & W-TALC\cite{paul2018w} & 54.5  & 61.5  & 50.5  &       & 1.0  & 0.2  & 0.0  & 0  & 0  & 0.2 \\
\cmidrule{1-5}\cmidrule{7-12}    \multirow{4}[2]{*}{Skeleton} & Zhu et.al\cite{zhu2021interpreting}   & 84.5(2.0) & 85.1(2.0) & 84.7(2.0) &       & \multicolumn{6}{c}{-} \\
          & MS-G3D\cite{liu2020disentangling} & 88.4(1.5) & 89.1(1.6) & 92.5(0.7) &       & \multicolumn{6}{c}{-} \\
          & STAM\cite{nguyen2021spatio}  & 86.5(2.6) & 86.5(3.0) & 93.1(1.8) &       & \multicolumn{6}{c}{-} \\
          & \textbf{WO-GMA} & \textbf{93.8}(1.0) & \textbf{94.4}(0.9) & \textbf{96.9}(0.7) &       & \textbf{31.7}  & \textbf{22.4}  & \textbf{17.9}  & \textbf{11.4}  & \textbf{5.2}   & \textbf{17.7} \\
    \bottomrule
    \end{tabular}%
  \label{tab1}%
\end{table}%

\textbf{Video-level classification performance}. For our model, one video is classified as F+ if the video level score got by Top-K strategy is greater than $0.5$ in the seen video. For STAM \cite{nguyen2021spatio}, MS-G3D \cite{liu2020disentangling}, Zhu et.al\cite{zhu2021interpreting}, the classification threshold is set to 0.5. Since action detection tasks also include classification, we also report the video-level performance of WOAD \cite{gao2021woad} and W-TALC \cite{paul2018w}. Following previous appearance-based methods \cite{gao2021woad,zhang2021cola}, the image features and optical features are extracted by I3D \cite{carreira2017quo} pre-trained on Kinetics \cite{kay2017kinetics}, a huge dataset contains 306k videos. For a fair comparison, none-skeleton based methods also use the first 6000 frames. Both the clip window size and window stride are 20. Other experiments settings are the same as original papers.

Results are shown in the left part of Table~\ref{tab1}. For skeleton-based methods, we report the 5-fold cross-validation results. For other methods, since the results perform much worse than skeleton-based methods and calculating image and optical flow features requires a lot of computing power, we only report results of the fifth fold with frame-level annotations. Though only accumulated history information is used, video-level results of our model outperform previous works by a lot margin, demonstrating the superiority of WO-GMA. Moreover, for both WOAD\cite{gao2021woad} and W-TALC \cite{paul2018w}, the model with the image input feature performs worse than the optical flow input feature. This result indicates that motion information is more important than appearance information in this task.

\begin{figure}[htbp]
\centering
\begin{minipage}[t]{0.46\textwidth}
\centering
\includegraphics[width=5cm]{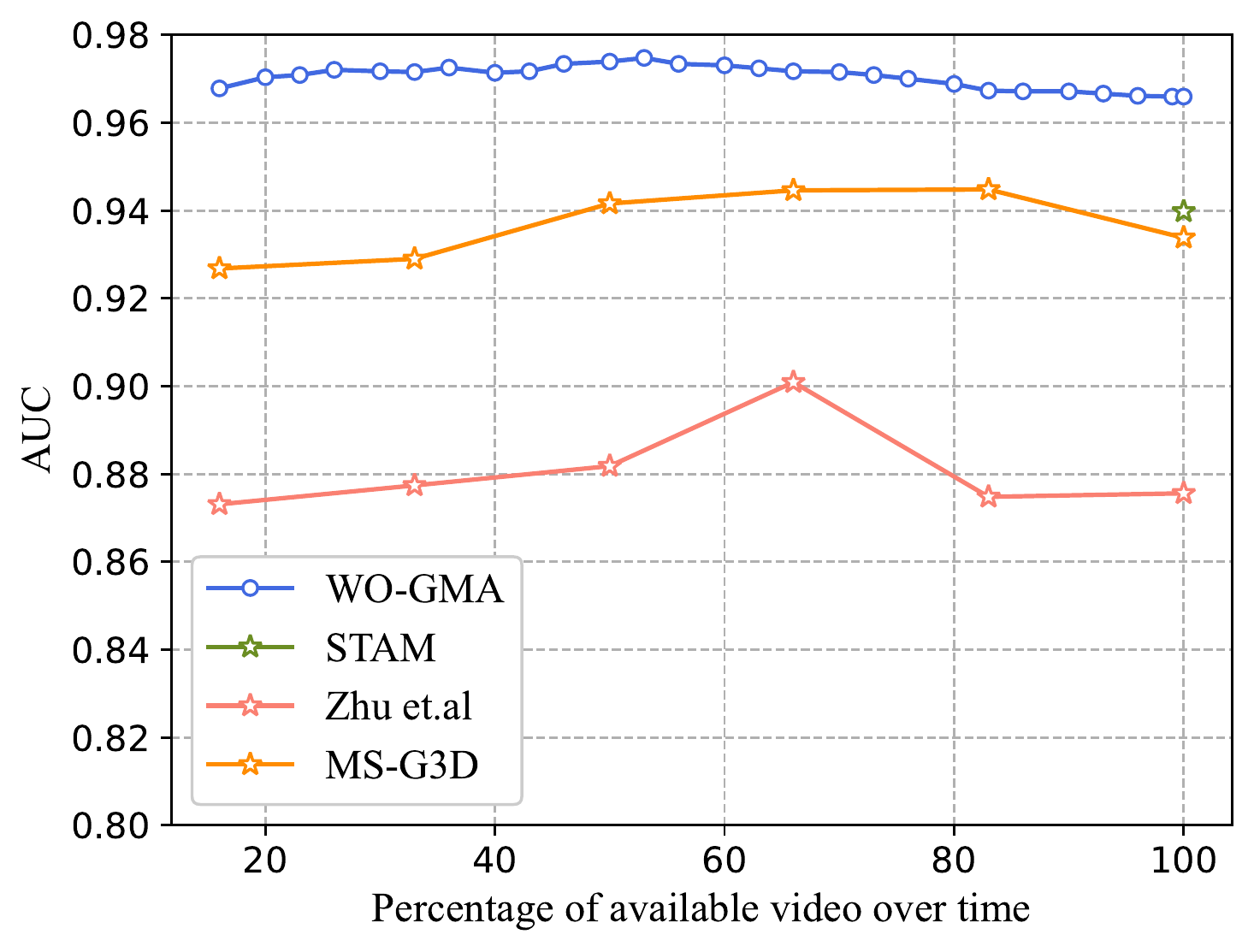}
\caption{Video-level performance comparison of AUC on our datasets when only part of frames is considered.}
\label{fig2}
\end{minipage}
\begin{minipage}[t]{0.46\textwidth}
\centering
\includegraphics[width=5cm]{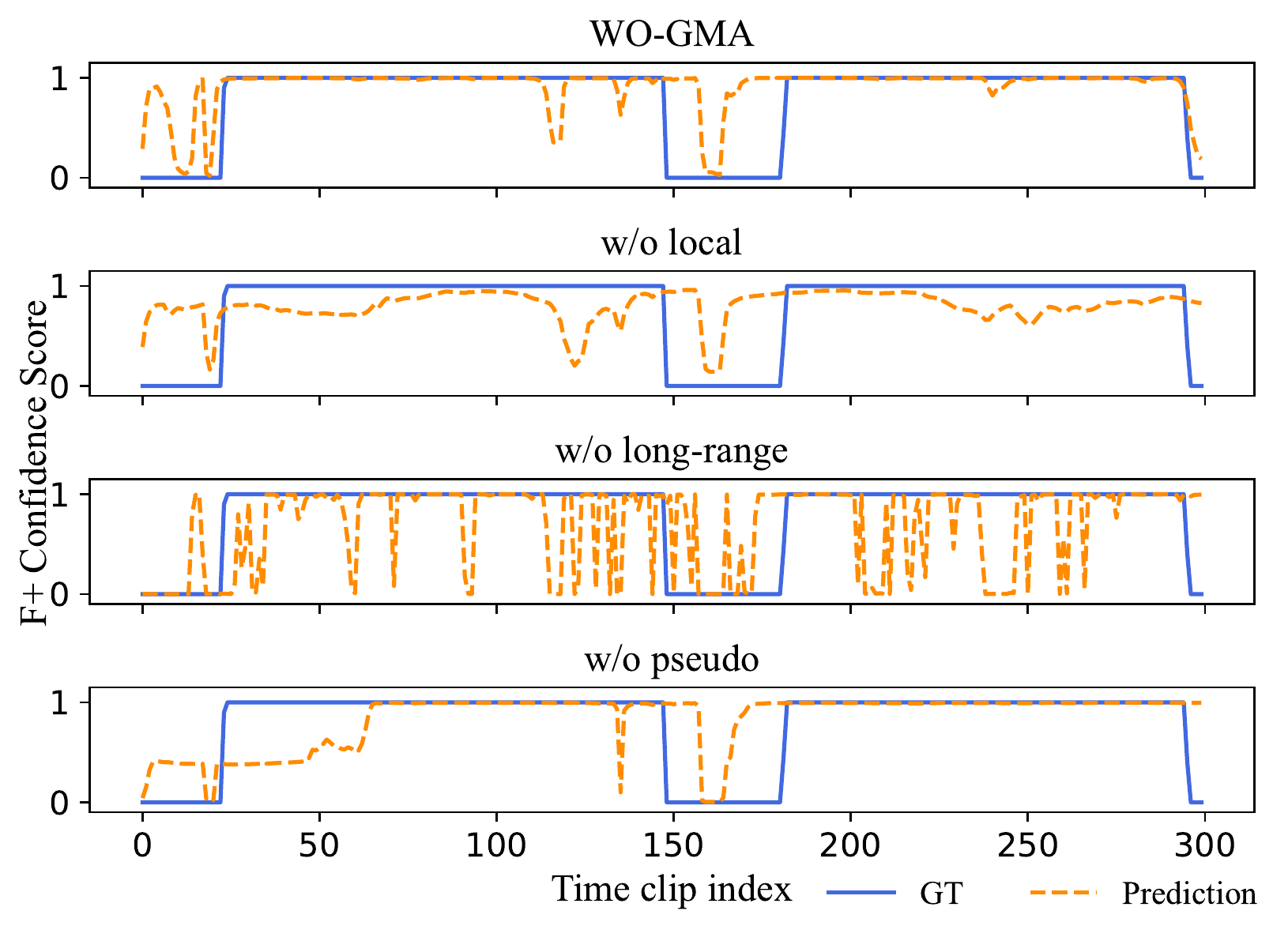}
\caption{An example of the comparison between detection results and ground truth frame-level annotations.}
\label{fig3}
\end{minipage}
\end{figure}

\textbf{Online action detection performance}. To the best of our knowledge, we are the first to develop weakly supervised online action detection method for GMA. 60 F+ samples in the fifth dataset split-fold were annotated by experts with frame-level labels, and 56 valid samples are used as ground truth to report detection results. The right part of Table~\ref{tab1} shows that our model achieves the best detection performance. Compared with the image features, the detection results using the optical flow features are better, which further demonstrates the importance of motion information. Since the skeleton contains only motion information, this result also proves the necessity of using the skeleton as input. The performance gap between left part and right part of Table~\ref{tab1} shows that detection is much more difficult than classification in this task. There are two main possible reasons. First, only video-level supervision information may not enough. Second, compared with everyday life actions like shaking hands, the boundary of FMs is even harder to determine for annotators accurately. Figure~\ref{fig2} demonstrates that our model can get video-level performance as good as fully observed when only the first 20\% video frames are observed. This result shows a significant benefit of our model: the assessment time of automated GMA for real-world applications can be greatly shortened. The visualisation detection results in the top subplot of Figure~\ref{fig3} shows the acceptable detection performance.

% \begin{figure}[t]
% 	\centering    %居中
% 	\subfigure[Local clip-level features] %
% 	{
% 		\begin{minipage}{0.4 \textwidth}
% 			%\centering      %
% 			\includegraphics[width=\textwidth]{figure/localT_SNE.png} 
% 		\end{minipage}
% 	}
% 	\subfigure[LSTM clip-level features] %
% 	{
% 		\begin{minipage}{0.4 \textwidth}
%   			\includegraphics[width=\textwidth]{figure/LSTMT_SNE.png}
% 		\end{minipage}
% 	}
% 	\caption{The t-SNE feature visualization of 20000 clip-level features, where red dots denote F+ clips and blue dots denote F- clips.} 
% 	\label{fig:reconstrSNR}  %
% \end{figure}

\subsection{Ablation Study}
Since the complex movements pattern of FMs, we argue that both local spatio-temporal information and long-range information are critical in this task. This part will analyze the effect each component in the fifth dataset split-fold with frame-level annotations.

To study the effect of CPGB which can combine future information, we remove this branch(w/o pseudo). As shown in Table~\ref{tab2}, both the classification and detection performance will drop compared with WO-GMA, demonstrating the necessity to generate pseudo clip labels for training. We replace the LFEM with only clip skeletons vertices features concatenation to get results without complex local features(w/o local). As shown in the second row of Table~\ref{tab2}, the video level accuracy will drop by $2\%$, and the detection results will drop when IoU is higher. Furthermore, to analyze the effect of long-range information, we remove the 1D convolutions used to capture long-range information in CPGB(w/o long-range). Results in Table~\ref{tab2} imply the importance of long-range information.

To better illustrate the influence of different modules, we plot three more curves of the same infant in Fig.3 and report the number of detection instances in Table~\ref{tab2}. These results show that the detection action instances are fragmented without long-range information, which is unsuitable for detecting continuous FMs. Without local feature extraction, the detection score is less confident than WO-GMA. Without pseudo, the model may ignore the gap between intermittent FMs, which the long-range information in CPGB will make up. Moreover, generating pseudo labels without long-range information will bring the noise. The detection results further show the difficulty of detection task mentioned in previous subsection. Compared with appearance-based methods, our skeleton-based methods achieve better performance.

\begin{table}[htbp]
  \centering
  \caption{Ablation analysis of our proposed WO-GMA}
    \begin{tabular}{lccccrrrrrrr}
    \hline
    \multirow{2}[2]{*}{Method} & \multicolumn{3}{c}{Video level} &       & \multicolumn{6}{c}{Detection – mAP@IoU(\%)}   & \multicolumn{1}{c}{\multirow{2}[2]{*}{Instances}} \bigstrut[t]\\
          & Accuracy & F1-score & AUC   &       & \multicolumn{1}{c}{0.1} & \multicolumn{1}{c}{0.2} & \multicolumn{1}{c}{0.3} & \multicolumn{1}{c}{0.4} & \multicolumn{1}{c}{0.5} & \multicolumn{1}{c}{mean} &  \bigstrut[b]\\
\cline{1-4}\cline{6-12}    w/o pseudo & 92.2  & 92.9  & 95.7  &       & 35.1  & 21.6  & 14.7  & 8.0     & 2.4   & 16.4  & 260 \bigstrut[t]\\
    w/o local & 92.8  & 93.6  & 95.8  &       & \textbf{33.7}  & \textbf{24.6}  & \textbf{19.4}  & 10.6  & 4.7   & 18.6  & 296 \\
    w/o long-range & 93.5  & 93.9  & 95.4  &       & 10.2  & 7.0     & 2.9   & 1.4   & 0.6   & 4.4   & 1000 \\
    our model & \textbf{94.8}  & \textbf{95.1}  & \textbf{96.6}  &       & 31.7  & 22.4  & 17.9  & \textbf{11.4}  & \textbf{5.2}   & 17.7  & 420 \bigstrut[b]\\
    \hline
    \end{tabular}%
  \label{tab2}%
\end{table}%

\section{Conclusion}
We are the first to propose WO-GMA to address online action detection for general movements assessment using weak supervision and evaluate it on a large dataset. Unlike previous methods that only focus on video classification, our WO-GMA can detect the occurrence of FMs in an online fashion without frame-level labels. Experiments results demonstrate that WO-GMA significantly outperforms state-of-the-art both in the classification and detection tasks. 

\bibliographystyle{splncs04}
\bibliography{ref}




% LNCS Homepage, \url{http://www.springer.com/lncs}. Last accessed 4
% \end{thebibliography}
\end{document}